\documentclass[journal, letterpaper]{IEEEtran}

\usepackage[T1]{fontenc}
\usepackage{textcomp}
\usepackage{graphicx}
\usepackage{url}        
\usepackage{amsmath}    
\usepackage{authblk}

\usepackage{textgreek}	
\usepackage{listings}
\usepackage{longtable}
\usepackage{array}
\usepackage{hyperref}
\usepackage{tikz}
\usetikzlibrary{shapes.multipart,shapes,arrows,positioning,fit,backgrounds}

\usepackage[
    style=numeric,
    sorting=none,
    defernumbers=false,
    citestyle=numeric,
    bibstyle=numeric
]{biblatex}
\DeclareFieldFormat{doi}{}
\DeclareFieldFormat{url}{}
\newcommand{\mytableref}[1]{Table \ref{#1}}

\title{AutoLoop: Fast Visual SLAM Fine-tuning through Agentic Curriculum Learning}
\author{Assaf Lahiany\textsuperscript{1}, Oren Gal\textsuperscript{1}}
\vspace{20pt}  
\affil{\textsuperscript{1}Swarm \& AI Lab (SAIL)\\Hatter Department of Marine Technologies\\Leon H. Charney School of Marine Sciences\\University of Haifa}

\begin{document}


\maketitle

\begin{abstract}
	Current visual SLAM systems face significant challenges in balancing computational efficiency with robust loop closure handling. Traditional approaches require careful manual tuning and incur substantial computational overhead, while learning-based methods either lack explicit loop closure capabilities or implement them through computationally expensive methods. We present AutoLoop, a novel approach that combines automated curriculum learning with efficient fine-tuning for visual SLAM systems. Our method employs a DDPG (Deep Deterministic Policy Gradient) agent to dynamically adjust loop closure weights during training, eliminating the need for manual hyperparameter search while significantly reducing the required training steps. The approach pre-computes potential loop closure pairs offline and leverages them through an agent-guided curriculum, allowing the model to adapt efficiently to new scenarios. Experiments conducted on TartanAir for training and validated across multiple benchmarks including KITTI, EuRoC, ICL-NUIM and TUM RGB-D demonstrate that AutoLoop achieves comparable or superior performance while reducing training time by an order of magnitude compared to traditional approaches. AutoLoop provides a practical solution for rapid adaptation of visual SLAM systems, automating the weight tuning process that traditionally requires multiple manual iterations. Our results show that this automated curriculum strategy not only accelerates training but also maintains or improves the model's performance across diverse environmental conditions.
\end{abstract}


\section{Introduction}

\IEEEPARstart{V}{isual} SLAM systems have become fundamental components in autonomous navigation and robotics applications, with recent learning-based approaches demonstrating impressive performance in challenging scenarios. However, integrating loop closure capabilities into these learned systems typically requires extensive retraining or complex architectural modifications. This is particularly evident in state-of-the-art methods like DPVO, which excel at frame-to-frame tracking but lack explicit loop closure handling.
We present AutoLoop, an efficient approach for enhancing learned visual SLAM systems with loop closure capabilities through automated curriculum fine-tuning. Our method leverages pre-computed loop closure pairs and a DDPG-based curriculum learning agent to rapidly adapt existing models while maintaining their core performance. By combining offline loop detection with automated weight adjustment, we achieve significant reductions in training time and computational requirements compared to traditional approaches.
The key contributions of our work include:
\begin{itemize}
	\item An efficient pre-computation pipeline for identifying and verifying loop closure pairs using a hybrid NetVLAD-SIFT approach.
	\item A DDPG-based curriculum learning strategy that automatically adjusts loop closure loss weights during fine-tuning.
	\item A targeted fine-tuning approach that reduces training steps by an order of magnitude while maintaining performance.
	\item Comprehensive evaluation on standard benchmarks demonstrating improved loop closure handling with minimal computational overhead.
\end{itemize}

This work addresses a critical gap in learning-based SLAM systems by providing an efficient and automated method for incorporating loop closure capabilities, making advanced visual SLAM more accessible for real-world applications.

\section{Background}

	Traditional visual SLAM systems like ORB-SLAM3 \cite{campos2021orb} and VINS-Mono \cite{qin2018vins} rely on hand-crafted features and optimization techniques for pose estimation and loop closure. Recent learning-based approaches, including DPVO \cite{lipson2024deep} and DeepV2D \cite{teed2018deepv2d}, have demonstrated superior performance in challenging scenarios by leveraging deep neural networks for feature extraction and matching. However, these learned systems often lack explicit loop closure handling, focusing primarily on frame-to-frame tracking accuracy. While DPV-SLAM \cite{lipson2025deep} extends DPVO with proximity-based loop closure detection, it incurs computational overhead and additional memory requirements due to its keypoint detector.

	Loop closure detection has been extensively studied in classical SLAM systems. DBoW2 \cite{zhang2019improved} and VLAD-based approaches \cite{huang2016vlad} have been widely adopted for place recognition, while NetVLAD \cite{arandjelovic2016netvlad} introduced learned descriptors for improved robustness. Recent works like SuperGlue \cite{sarlin2020superglue} and LoFTR \cite{sun2021loftr} have explored learned feature matching for geometric verification, though their integration into end-to-end SLAM systems remains challenging due to computational constraints.

	Curriculum learning \cite{bengio2009curriculum}, \cite{weinshall2018curriculum} has shown promise in various computer vision tasks, including SLAM. Works like gradSLAM \cite{jatavallabhula2019gradslam} and DeepFactors \cite{czarnowski2020deepfactors} introduced manual curriculum strategies for training deep SLAM systems. However, these approaches typically require careful hand-tuning of learning schedules and lack automation in curriculum design.

	Recent developments in efficient model adaptation, such as LoRA \cite{devalal2018lora} and prompt tuning \cite{jia2022visual}, have demonstrated the benefits of targeted fine-tuning over full model retraining. In the context of SLAM, works like TC-SLAM \cite{gopaul2018optimal} and DeepLIO \cite{iwaszczuk2021deeplio} have explored transfer learning for domain adaptation, though they typically require substantial training data and time.

	Pre-computation strategies have been explored in various SLAM contexts, primarily for map building and relocalization. Systems like PTAM \cite{pire2017s} and MapLab \cite{schneider2018maplab} utilize pre-built maps for efficient localization, while recent works like Hydra \cite{udugama2023mono} and AtLoc \cite{wang2020atloc} leverage pre-computed features for improved performance. Our work extends this concept to loop closure detection, using pre-computed pairs for efficient training supervision.

	Our approach uniquely combines these elements - automated curriculum learning, efficient fine-tuning, and pre-computed loop closures - to address the specific challenge of integrating loop closure capabilities into learned visual SLAM systems. Unlike previous works that either require extensive training or manual tuning, our method achieves efficient adaptation while maintaining real-time performance.

\section{Methodology}

	Our approach consists of three main components: (1) an efficient loop closure detection pipeline for pre-computing potential loop pairs, (2) a DDPG-based curriculum learning agent for automated weight adjustment, and (3) a fine-tuning strategy that leverages pre-computed loops and curriculum learning to rapidly adapt DPVO models. We build upon the DPVO architecture, which provides state-of-the-art performance in visual odometry through its dense feature extraction and matching capabilities. The system is primarily developed and validated on the TartanAir dataset, which is particularly suited for loop closure learning due to its rich variety of revisited locations under different viewing angles, lighting conditions, and seasonal changes. These naturally occurring loop closure scenarios, combined with accurate ground truth poses, provide an ideal training environment for our approach. For evaluation, we test on standard benchmarks including KITTI \cite{geiger2013vision}, EuRoC \cite{burri2016euroc}, ICL-NUIM \cite{handa2014benchmark} and TUM RGB-D \cite{sturm2012benchmark} datasets to demonstrate generalization capabilities.

\subsection{Off-Line Loop Closure Database}

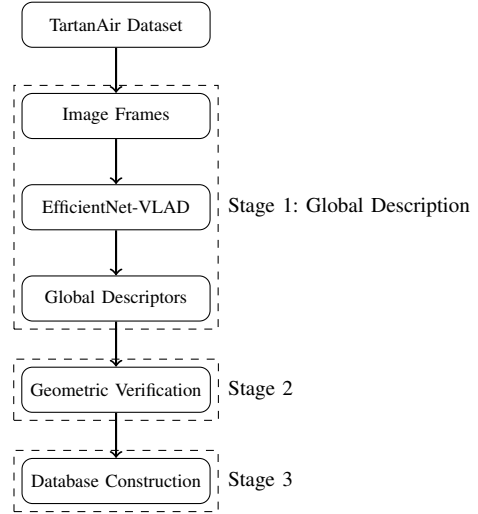
\begin{figure}[t]
    \centering
    \begin{tikzpicture}[
        scale=0.5,
        node distance=0.6cm,
        box/.style={rectangle, draw, minimum width=2.5cm, minimum height=0.6cm, text centered, rounded corners, font=\scriptsize},
        group/.style={rectangle, draw, dashed, inner sep=0.1cm},
        arrow/.style={->, thick}
    ]
        \node[box] (dataset) {TartanAir Dataset};
        
        \node[box, below=of dataset] (frames) {Image Frames};
        \node[box, below=of frames] (netvlad) {EfficientNet-VLAD};
        \node[box, below=of netvlad] (descriptors) {Global Descriptors};
        
        \node[box, below=of descriptors] (stage2) {Geometric Verification};
        \node[box, below=of stage2] (stage3) {Database Construction};
        
        \draw[arrow] (dataset) -- (frames);
        \draw[arrow] (frames) -- (netvlad);
        \draw[arrow] (netvlad) -- (descriptors);
        \draw[arrow] (descriptors) -- (stage2);
        \draw[arrow] (stage2) -- (stage3);
        
        \begin{pgfonlayer}{background}
            \node[group, fit=(frames) (netvlad) (descriptors), label=right:{\footnotesize Stage 1: Global Description}] {};
            \node[group, fit=(stage2), label=right:{\footnotesize Stage 2}] {};
            \node[group, fit=(stage3), label=right:{\footnotesize Stage 3}] {};
        \end{pgfonlayer}
    \end{tikzpicture}
    \caption{Loop Closure Detection Pipeline: The system processes TartanAir sequences through three stages to identify and validate loop closure pairs.}
    \label{fig:pipeline}
\end{figure}

	To create a reliable ground truth database for fine-tuning our visual odometry model with loop closure constraints we use construct an offline loop closure detection pipeline (Figure \ref{fig:pipeline}). The system processes visual sequences from the TartanAir dataset through a three-stage architecture to identify and validate loop closure pairs. First, a global descriptor module based on EfficientNet-VLAD generates compact representations of each frame, enabling efficient similarity-based retrieval of loop closure candidates. Second, a geometric verification stage validates these candidates using local feature matching and epipolar geometry constraints, ensuring spatial consistency. Finally, a database construction module aggregates the validated loop closures, storing frame pairs along with their confidence metrics in a structured format. 
	

	This offline approach allows for exhaustive loop closure detection without real-time constraints, creating a robust ground truth database. The resulting dataset serves as supervision during the fine-tuning phase of our visual odometry model, where loop closure constraints are incorporated into the training loss function to enhance the model's capability for globally consistent trajectory estimation. This decoupled architecture separates the computationally intensive loop closure detection from the training process while ensuring high-quality supervision for the learning task.
	
\subsection{Baseline Model}

As our baseline model architecture, we use the Deep Patch Visual Odometry (DPVO) model, introduced in \cite{lipson2024deep}. DPVO is particularly well-suited for our loop closure enhancement approach due to its modular loss structure, which can be readily extended to incorporate additional supervision signals. The model's existing decomposition into pose estimation and optical flow components provides a natural framework for integrating loop closure constraints without disrupting the core visual odometry capabilities. DPVO's end-to-end trainable nature allows us to smoothly incorporate the loop closure loss during fine-tuning while preserving the model's fundamental trajectory estimation abilities

\subsection{DPVO Loss Supervision} The DPVO loss function is carefully constructed to balance the contributions of pose estimation and optical flow prediction through weighted supervision terms. The total loss comprises two primary components: a pose loss (\(\mathcal{L}_{pose}\)) that supervises the relative camera motion estimation, and a flow loss (\(\mathcal{L}_{flow}\)) that guides the prediction of dense optical flow fields between consecutive frames. Given the different numerical scales and impact of these terms, appropriate scaling weights are crucial for stable training and optimal performance. 
	
	\begin{equation}
		\mathcal{L}_{total} = s_{p}\mathcal{L}_{pose} + s_{f}\mathcal{L}_{flow}
		\label{eq:totalLoss}
	\end{equation}
		
	As established in \cite{lipson2024deep} the pose loss is scaled by a factor of $s_{p}=10$ to emphasize the importance of accurate camera trajectory estimation, while the flow loss is scaled by $s_{f}=0.1$ to prevent the optical flow supervision from dominating the learning process. This empirically determined weighting scheme (\ref{eq:totalLoss}) ensures that both components contribute meaningfully to the network's learning objective while maintaining their relative importance in the overall optimization landscape.

\subsection{Loop Closure Aware Loss}

    Our methodology modifies DPVO's training objective by adding a loop closure loss component to the original loss function. This additional component enables the model to learn from loop closure constraints while maintaining its core visual odometry capabilities.
	
	\begin{equation}
		\mathcal{L}_{total} = s_{f}\mathcal{L}_{flow} + s_{p}\mathcal{L}_{pose} + w_{loop}\mathcal{L}_{loop}
		\label{eq:CLtotalLoss}
	\end{equation}
		
	\(s_{f}, s_{p}\) are the scaling factors introduced in the original DPVO loss function (\ref{eq:totalLoss}).
	where \(w_{loop}\) is the weight of the loop closure loss component. Definition of the \(w_{loop}\) is critical to the performance of the overall training process. We define loop closure loss as the sum of the relative transformation errors between each predicted pose and its corresponding pre-calculated pairs. For a given frame's predicted pose \(T_{\text{pred},i}\), the loss aggregates the relative transformation errors with all its valid pre-calculated loop closure pairs \(G_{\text{gt},j}\).

	\begin{equation}
		\mathcal{L}_{loop} = \frac{1}{N} \sum_{i,j}^{N} h_{\delta}(\|Log_{SE(3)}(T_{\text{pred},i}^{-1} \cdot G_{\text{gt},j})\|)
	\label{eq:loopLoss}
	\end{equation}

	where Where $N$ is the total number of valid loop closure pairs and $h_{\delta}(x)$ is the Huber loss function defined as:

	\begin{equation}
	h_{\delta}(x) = \begin{cases}
	\frac{1}{2}x^2 & \text{if } |x| \leq \delta \\
	\delta|x| - \frac{1}{2}\delta^2 & \text{otherwise}
	\end{cases}
	\label{eq:huberLoss}
	\end{equation}

	The huber loss addresses two critical aspects: providing precise gradients for fine-tuned corrections when errors are small, while preventing gradient explosion from potentially incorrect loop closure pairs or challenging viewpoint changes when errors are large. This balanced approach ensures stable and effective learning from geometric constraints while remaining robust to noise in the loop closure pairs.
	
\subsection{Agentic Curriculum-Learning}

	The DPVO framework employs an adaptive curriculum learning strategy using a Deep Deterministic Policy Gradient (DDPG) agent to dynamically adjust the loop closure loss weight during training. Unlike traditional fixed or manually scheduled weights, the DDPG agent learns to optimize the loop closure weight based on the training dynamics and current model performance. The agent observes the current loss values and training progress as its state, and outputs a continuous action representing the loop weight adjustment. This approach is particularly valuable for loop closure supervision because the importance of loop closure constraints can vary significantly depending on the training stage and scene characteristics. 

    \begin{equation}
    	w_{i}^{loop} = w_{0} + (w_{F} - w_{0})a_{i}
	\label{eq:adaptiveWeights}
	\end{equation}
	\vspace{-12pt}
	\begin{equation}
		\mathcal{L}_{i}^{ema} = \alpha\mathcal{L}_{i-1}^{ema} + (1-\alpha)|\mathcal{L}_{i}^{loop}|
	\label{eq:emaLoss}
	\end{equation}
	\vspace{-12pt}
	\begin{equation}
    	s_{i} = [p_{i},\ \mathcal{L}_{i}^{ema}]; \ \ r_{i} = - \mathcal{L}_{i}^{ema}
	\label{eq:adaptiveState}
	\end{equation}
	\vspace{-12pt}
	\begin{equation}
    	a_{i} = \mu_{k}(s_{i}) + \mathcal{N}_{i}
	\label{eq:adaptiveAction}
	\end{equation}
	
	The DDPG agent in this code operates on a 2-dimensional state space that combines the training progress $p_{i} \in [0,1]$ and the smoothed loss value $\mathcal{L}_{i}^{ema}$. For each training step, it outputs a single action value that determines how to interpolate between initial and final weights (\ref{eq:adaptiveWeights}). The agent learns from experience by storing transitions of (state, action, reward, next state) tuples, where the reward $r_{i}$ is simply the negative of the smoothed loss. This setup allows the agent to adaptively adjust curriculum weights based on both the training progress and current performance, effectively learning an optimal progression path that minimizes loop closure loss during training.

	This adaptive weighting is particularly crucial during fine-tuning, as it helps prevent catastrophic forgetting of the original model's capabilities while gradually introducing loop closure supervision. The DDPG agent learns to modulate the importance of loop closure constraints based on the model's current adaptation state, ensuring that the geometric consistency is enhanced without compromising the fundamental pose estimation accuracy developed in the original training.

\subsection{Loop Closure Aware Fine-Tuning}

	We leverage the pre-computed loop closure pairs to fine-tune the DPVO model. During training, we specifically focus on sequences containing verified loop closures, ensuring efficient learning of loop closure handling. Our sampling strategy selects trajectories where our pre-computation pipeline has successfully identified loop pairs, as these sequences provide the necessary supervision for both standard pose estimation and loop closure scenarios. This targeted approach ensures that each training iteration contributes meaningfully to enhancing the model's loop closure capabilities while maintaining its core visual odometry performance. The DDPG agent dynamically modulates the loop closure weight $w_{loop}$ (\ref{eq:adaptiveWeights}), typically initializing at conservative values (0.1-0.2) and progressively increasing based on model stability and performance. This pre-computation strategy offers multiple benefits: it eliminates the computational burden of online loop detection during training, ensures consistent supervision through verified loop closure pairs, and enables efficient batch processing of loop constraints.

	Consequently, our approach strive to achieve effective integration of loop closure capabilities into the DPVO framework with high efficiency and autonmous agentic weight tuning, requiring only a fraction of the training time and resources needed for training from scratch.

\section{Experiments}

	We evaluate our methodology on EuRoC MAV \cite{burri2016euroc}, KITTI \cite{geiger2013vision} odometry benchmark, TUM RGB-D \cite{sturm2012benchmark}, ICL-NUIM \cite{handa2014benchmark} dataset and TartanAir test set from ECCV 2020 SLAM competition . Each experiment is run 5 times and we report the median result. We compare our AutoLoop method to both pure VO and SLAM variants. Given AutoLoop's VO-based architecture, it offers dual functionality: it can operate as a pure VO model, potentially serving as a drop-in replacement in existing SLAM pipelines, or function as a standalone system with loop closure capabilities. All experiments were conducted on an NVIDIA DGX-1 computing node equipped with 8 V100 GPUs, enabling parallel processing and rapid validation across our methodological variants.

\subsection{Pre-compute Loop Closure Pairs}
	Our loop closure pipeline implementation employs several critical parameters that govern its loop closure detection behavior. The system maintains a circular buffer of the most recent 2000 frames for potential loop closure candidates. To identify revisited locations, we utilize a similarity metric with a threshold of 0.75, where higher values indicate stronger matches between frame descriptors. The NetVLAD descriptor is configured with 32 cluster centers, offering a good trade-off between descriptor discriminability and computational overhead. For robust geometric verification, we require a minimum of 30 inlier feature matches between candidate frame pairs to confirm a valid loop closure, effectively filtering out false positives while maintaining high recall.

\begin{figure}[ht!]
			\begin{center}
				\includegraphics[scale=0.2]{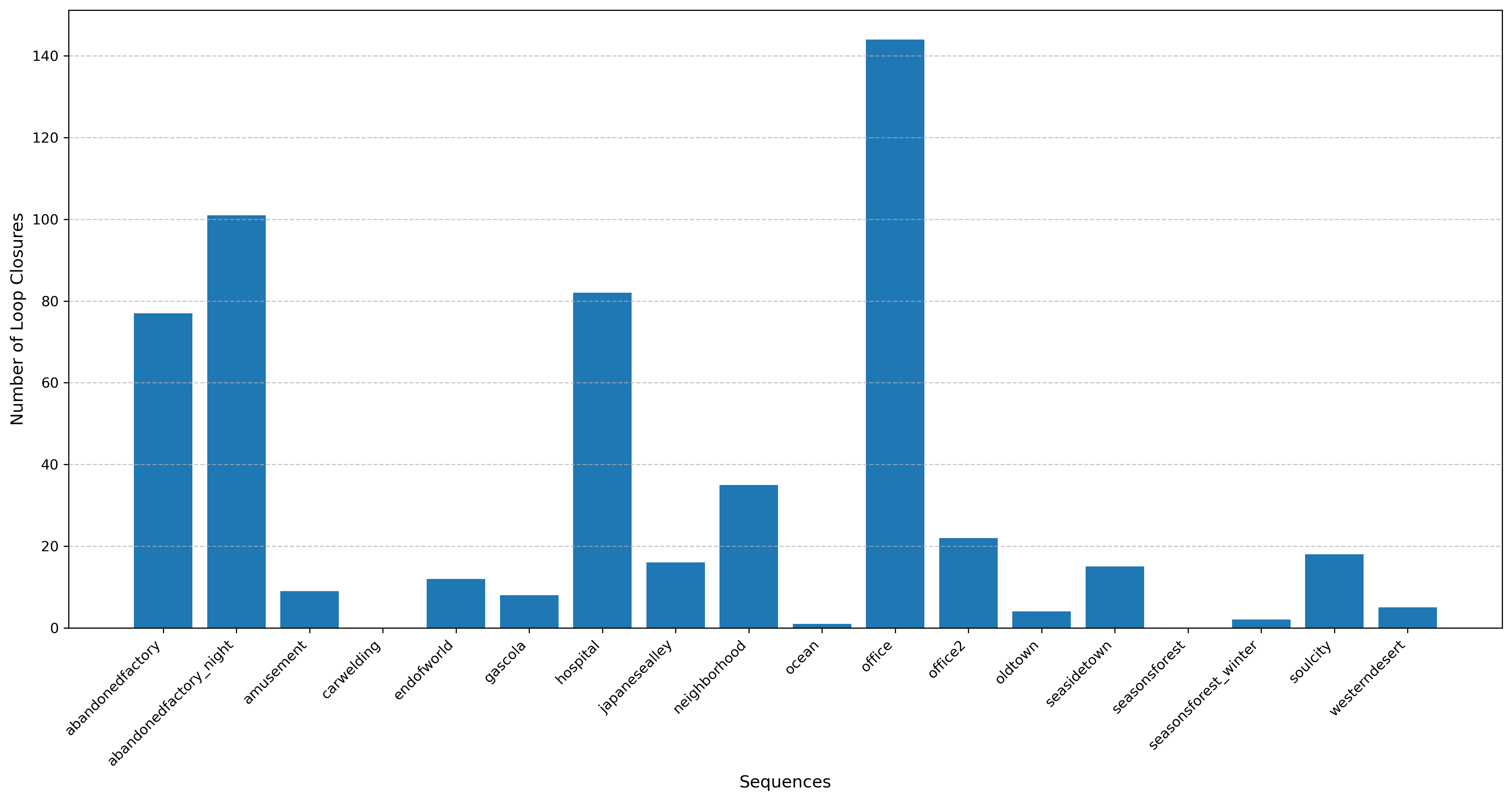}
				\caption{Pre-computed loop closure pairs distribution on TartanAir training set. A total of 337 scenes processed with 551 loop closure pairs detected.}
				\label{fig:loopClosureHistogram}
		\end{center}
	\end{figure}

	The offline pre-compute on TartanAir produced 551 loop closure pairs. Figure \ref{fig:loopClosureHistogram} shows the distribution of loop closure pairs across the TartanAir training set.
		
\subsection{Fine Tuning}
	During DPVO fine tuning we use only those sequences that contain loop closure pairs. The sampled trajectories includes both frames with loop closure pairs and frames without, ensuring all loss signals (pose, flow, loop) are present.
    
\begin{figure}[ht!]
        \begin{center}
            \includegraphics[scale=0.2]{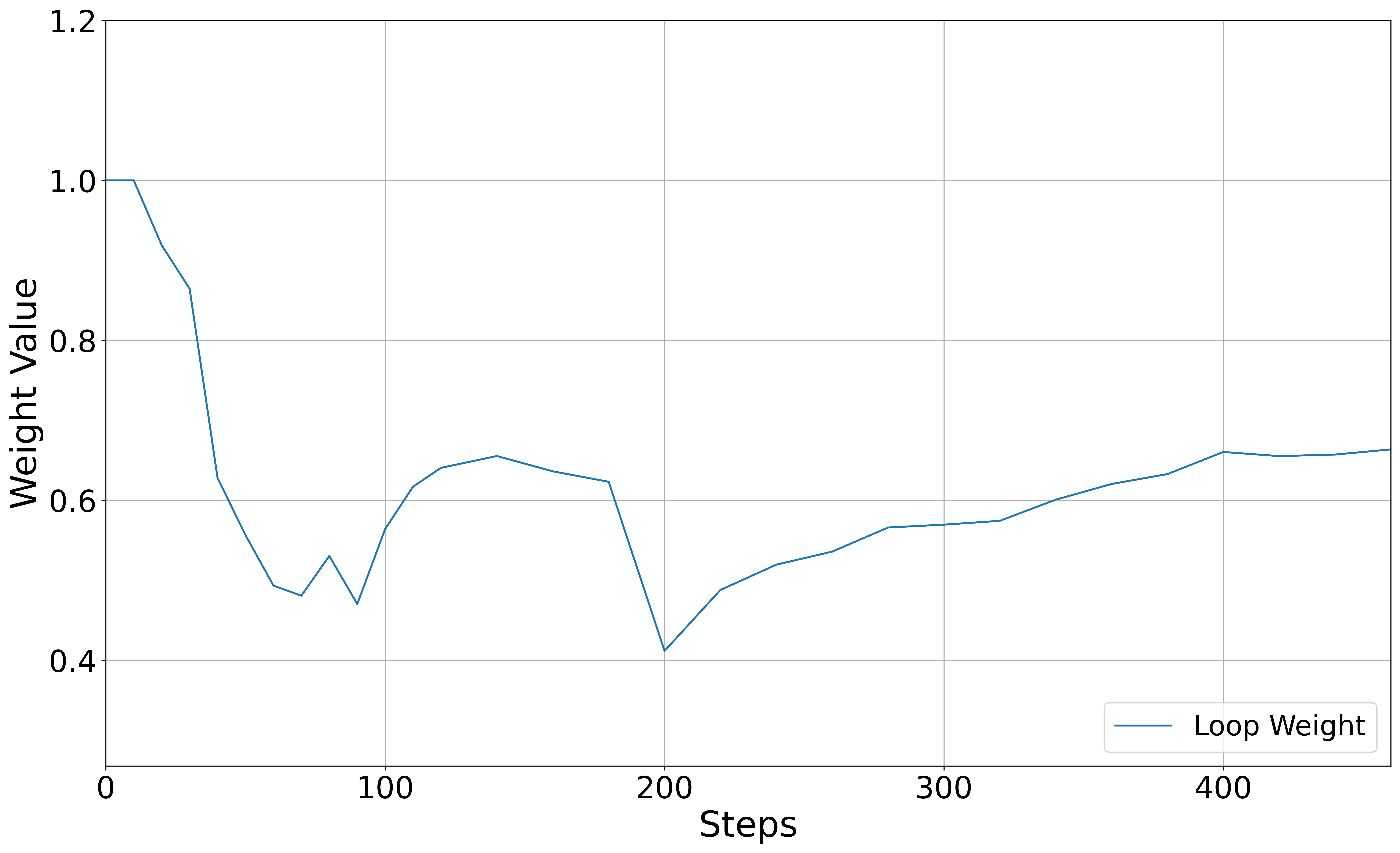}
            \caption{Progression of the loop weight during DPVO fine-tuning using the incorporated loop closure loss variant.}
            \label{fig:adaptiveLearningCLTrain}
	\end{center}
    \end{figure}

	\vspace{0.5em}
	\noindent
    \textbf{Adaptive Weighting}: Our RL DDPG agent is structured as in \cite{lahiany2024robust} with three-layer actor/critic networks (max width of 64). Since we use smaller dataset we reduce the agent training frequency to every 30 global DPVO training steps compared the CL-DPVO-RL-DDPG approach \cite{lahiany2024robust}, using batch size of 64 samples from a 5k-sized replay buffer. Figure \ref{fig:adaptiveLearningCLTrain} shows the loop weight progression during fine tuning. A short exploration stage (till step 200) is followed by convergance to 0.62 as the model start overfitting after step 420.
	
	\begin{table*}[!t]
		\begin{center}
		\caption{Results on the KITTI Monocular SLAM training set. AutoLoop outperforms both DPVO and DPV-SLAM while running at superior FPS as DPVO VO model.}
		\renewcommand{\arraystretch}{1.2}
		\label{tab:kittiAte}
		\begin{tabular}{|c|c@{\hspace{3pt}}c@{\hspace{3pt}}c@{\hspace{3pt}}c@{\hspace{3pt}}c@{\hspace{3pt}}c@{\hspace{3pt}}c@{\hspace{3pt}}c@{\hspace{3pt}}c@{\hspace{3pt}}c@{\hspace{3pt}}c@{\hspace{3pt}}c|c|}
			\hline
			\hline
			& 00 & 01 & 02 & 03 & 04 & 05 & 06 & 07 & 08 & 09 & 10 & Avg & FPS \\
			\hline
			ORB-SLAM2* & 8.27 & X & \textbf{26.86} & \textbf{1.21} & 0.77 & 7.91 & 12.54 & 3.44 & \textbf{46.81} & 76.54 & \textbf{6.61} & - & 34 \\
			ORB-SLAM3* & \textbf{6.77} & X & 30.500 & 1.036 & 0.930 & 5.542 & 16.605 & 9.700 & 60.687 & 7.899 & 8.650 & - & 34 \\
			LDSO & 9.32 & 11.68 & 31.98 & 2.85 & 1.22 & \textbf{5.1} & 13.55 & 2.96 & 129.02 & \textbf{21.64} & 17.36 & 22.42 & 49 \\
			DROID-VO & 98.43 & 84.2 & 108.8 & 2.58 & 0.93 & 59.27 & 64.4 & 24.2 & 64.55 & 71.8 & 16.91 & 54.19 & 17 \\
			DPVO & 113.21 & 12.69 & 123.4 & 2.09 & 0.68 & 58.96 & 54.78 & 19.26 & 115.9 & 75.1 & 13.63 & 53.61 & 48 \\
			DROID-SLAM* \cite{teed2021droid} & 92.1 & 344.6 & X & 2.38 & 1.00 & 118.5 & 62.47 & 21.78 & 161.6 & X & 118.7 & - & 17 \\
			DPV-SLAM* \cite{lipson2025deep} & 112.8 & 11.50 & 123.53 & 2.50 & 0.81 & 57.80 & 54.86 & 18.77 & 110.49 & 76.66 & 13.65 & 53.03 & 39 \\
			DPV-SLAM++* & 8.30 & 11.86 & 39.64 & 2.50 & 0.78 & 5.74 & \textbf{11.60} & \textbf{1.52} & 110.9 & 76.70 & 13.70 & \textbf{25.76} & 39 \\
			\hline
			AutoLoop & 111.11 & \textbf{11.11} & 111.33 & 2.38 & \textbf{0.60} & 55.72 & 55.89 & 18.79 & 105.79 & 74.78 & 12.95 & 50.95 & 48 \\
			\hline
		\end{tabular}
		\end{center}
	\end{table*}

	\begin{table*}[!t]
    \begin{center}
    \caption{Results on the TartanAir monocular test split from the ECCV 2020 SLAM competition. Results are reported as ATE with scale alignment. Top performing method marked in bold. Methods marked with (*) use global optimization / loop closure.}
	\renewcommand{\arraystretch}{1.2}
	\label{tab:tartanAirTestSplitAte}
    \begin{tabular}{|c|c@{\hspace{3pt}}c@{\hspace{3pt}}c@{\hspace{3pt}}c@{\hspace{3pt}}c@{\hspace{3pt}}c@{\hspace{3pt}}c@{\hspace{3pt}}c@{\hspace{3pt}}c@{\hspace{3pt}}c@{\hspace{3pt}}c@{\hspace{3pt}}c@{\hspace{3pt}}c@{\hspace{3pt}}c@{\hspace{3pt}}c@{\hspace{3pt}}c@{\hspace{3pt}}|c@{\hspace{3pt}}c@{\hspace{3pt}}|}
		\hline
        \hline
        & ME & ME & ME & ME & ME & ME & ME & ME & MH & MH & MH & MH & MH & MH & MH & MH & Avg & Avg\\
        & 000 & 001 & 002 & 003 & 004 & 005 & 006 & 007 & 000 & 001 & 002 & 003 & 004 & 005 & 006 & 007 & Total & MH \\
		\hline
		ORB-SLAM3* \cite{campos2021orb} & 13.61 & 16.86 & 20.57 & 16.00 & 22.27 & 9.28 & 21.61 & 7.74 & 14.44 & 2.92 & 13.51 & 8.18 & 2.59 & 21.91 & 11.70 & 25.88 & 14.38 & 12.64 \\
		DROID-SLAM* \cite{teed2021droid} & 0.17 & 0.06 & 0.36 & 0.87 & 1.14 & 0.13 & 1.13 & \textbf{0.06} & \textbf{0.08} & 0.05 & \textbf{0.04} & \textbf{0.02} & \textbf{0.01} & 0.68 & 0.30 & 0.07 & 0.33 & 0.24 \\
		DROID-VO \cite{teed2021droid} & 0.22 & 0.15 & 0.24 & 1.27 & 1.04 & 0.14 & 1.32 & 0.77 & 0.32 & 0.13 & 0.08 & 0.09 & 1.52 & 0.69 & 0.39 & 0.97 & 0.58 & 0.52 \\
		DPVO \cite{lipson2024deep} & 0.16 & 0.11 & 0.11 & 0.66 & 0.31 & 0.14 & 0.30 & 0.13 & 0.21 & 0.04 & \textbf{0.04} & 0.08 & 0.58 & \textbf{0.17} & 0.11 & 0.15 & 0.21 & 0.17\\
		RAMP-VO \cite{pellerito2024deep} & 0.20 & \textbf{0.04} & \textbf{0.10} & 0.46 & \textbf{0.16} & 0.13 & \textbf{0.12} & 0.12 & 0.36 & 0.06 & \textbf{0.04} & 0.04 & 0.41 & 0.25 & 0.11 & \textbf{0.07} & 0.17 & 0.17 \\
		CL-DPVO \cite{lahiany2024robust} & \textbf{0.10} & 0.05 & 0.14 & 0.38 & 0.19 & \textbf{0.06} & 0.34 & 0.11 & 0.26 & \textbf{0.03} & 0.05 & \textbf{0.02} & 0.18 & 0.21 & 0.10 & 0.10 & \textbf{0.14} & \textbf{0.12} \\
        DPV-SLAM* \cite{lipson2025deep} & - & - & - & - & - & - & - & - & 0.23 & 0.05 & \textbf{0.04} & 0.04 & 0.54 & 0.15 & \textbf{0.07} & 0.14 & - & 0.16 \\
		DPV-SLAM++* & - & - & - & - & - & - & - & - & 0.21 & 0.04 & \textbf{0.04} & 0.04 & 0.92 & 0.17 & 0.11 & 0.13 & - & 0.21 \\
		\hline
        AutoLoop & 0.16 & 0.05 & 0.11 & \textbf{0.24} & 0.39 & 0.12 & 0.40 & 0.14 & 0.19 & 0.04 & 0.05 & 0.04 & 0.33 & 0.18 & 0.11 & \textbf{0.07} & 0.17 & 0.13 \\
		\hline
    \end{tabular}
    \end{center}
\end{table*}

\subsection{Benchmarks Comparison}

	\noindent
	\textbf{KITTI \cite{geiger2013vision}}: We evaluate our AutoLoop model on sequences 00-10 from KITTI training set. Table \mytableref{tab:kittiAte} shows improvement in avarage ATE on both DPVO and its proximity loop closure variant DPV-SLAM while still maintaining the high FPS of DPVO (visual odometry only). Compared to DPV-SLAM++, which is the DPVO based SLAM variat that combine both proximity and classic loop closure, we sucrifice ATE performance for a large increase in FPS. on sequences 01, 04 we show best overall performance.

	\vspace{0.5em}
	\noindent
    \textbf{TartanAir Test Split \cite{wang2020tartanair}}: We compare our AutoLoop models with state-of-the-art (SOTA) methods on the TartanAir test-split from the ECCV 2020 SLAM competition, including improved image and event mixture VO methods \cite{klenk2024deep}. The AutoLoop achives comparable results to the best in class CL-DPVO (0.13 vs 0.12) outperforming other SLAM methods such as DROID-SLAM and DPV-SLAM (0.24,0.16 vs 0.13) which incorporate the computational overhead of SLAM optimization and loop closure.
    
	\vspace{0.5em}
	\noindent
    \textbf{EuRoC MAV \cite{burri2016euroc}}: When comparing AutoLoop to both VO and SLAM methods on Machine-Hall and Vicon 1 \& 2 sequences from the EuRoC MAV dataset it struggle to generelaize well its loop closure capabilities to the indoor characteristics of the dataset. It underperforms the CL-DPVO and DPV-SLAM++ methods altho it does outperform the original DPVO.

	\vspace{1.5em}
	\noindent
    \textbf{TUM-RGBD \cite{sturm2012benchmark}}: In \mytableref{tab:TUM-RGBDTestSplitAte}, we benchmark AutoLoop on the Freiburg1 set of TUM-RGBD dataset. This benchmark is used to evaluate indoor SLAM performance and has challenging sequences with erratic camera movements and substantial motion blur. As in EuRoC MAV \cite{burri2016euroc}, AutoLoop struggles compared to the SLAM methods incapable of showing its loop closure capabilities.

	\begin{table}[!hbt]
		
		\begin{center}
			
			\caption{Results of the Avg. ATE[m] on the EuRoC test split monocular SLAM dataset}
			\label{tab:EuRoCTestSplitAte}
			\renewcommand{\arraystretch}{1.3}
			\begin{tabular}{|@{\hspace{3pt}}c@{\hspace{3pt}}|@{\hspace{3pt}}c@{\hspace{3pt}}@{\hspace{3pt}}c@{\hspace{3pt}}@{\hspace{3pt}}c@{\hspace{3pt}}@{\hspace{3pt}}c@{\hspace{3pt}}@{\hspace{3pt}}c@{\hspace{3pt}}@{\hspace{3pt}}c@{\hspace{3pt}}@{\hspace{3pt}}c@{\hspace{3pt}}@{\hspace{3pt}}|c@{\hspace{3pt}}|}
				\hline
				\hline
				& \rotatebox{90}{TartanVO \cite{wang2021tartanvo}} & \rotatebox{90}{SVO \cite{forster2014svo}} & \rotatebox{90}{DSO \cite{engel2017direct}} & \rotatebox{90}{DROID-VO \cite{teed2021droid}}& \rotatebox{90}{DPVO \cite{lipson2024deep}}& \rotatebox{90}{CL-DPVO \cite{lahiany2024robust}}& \rotatebox{90}{DPV-SLAM++ \cite{lipson2025deep} }& \rotatebox{90}{AutoLoop}\\
				\hline
				MH01 & 0.639 & 0.100 & \textbf{0.046} & 0.163 & 0.087 & 0.081 & 0.013 & 0.580  \\
				MH02 & 0.325 & 0.120 & 0.046 & 0.121 & 0.055 & \textbf{0.030} & 0.016 & 0.054 \\
				MH03 & 0.550 & 0.410 & 0.172 & 0.242 & 0.158 & 0.122 & 0.021 & 0.129 \\
				MH04 & 1.153 & 0.430 & 3.810 & 0.399 & 0.137 & \textbf{0.133} & 0.041 & 0.132 \\
				MH05 & 1.021 & 0.300 & \textbf{0.110} & 0.270 & 0.114 & 0.114 & 0.041 & 0.093 \\
				V101 & 0.447 & 0.070 & 0.089 & 0.103 & \textbf{0.050} & 0.051 & 0.035 & 0.047 \\
				V102 & 0.389 & 0.210 & \textbf{0.107} & 0.165 & 0.140 & 0.118 & 0.010 & 0.151 \\
				V103 & 0.622 & -	 & 0.903 & 0.158 & 0.086 & 0.063 & 0.015 & 0.094 \\
				V201 & 0.433 & 0.110 & \textbf{0.044} & 0.102 & 0.057 & 0.065 & 0.021 & 0.059 \\
				V202 & 0.749 & 0.110 & 0.132 & 0.115 & 0.049 & \textbf{0.045} & 0.011 & 0.046 \\
				V203 & 1.152 & 1.080 & 1.152 & 0.204 & 0.211 & \textbf{0.178} & 0.023 & 0.203 \\
				\hline
				Avg & 0.680 & 0.294 & 0.601 & 0.186 & 0.105 & \textbf{0.091} & 0.023 & 0.097 \\
				\hline
			\end{tabular}
		\end{center}
	\end{table}

	\begin{table}[!hbt]
    	
    	\begin{center}
    		
    		\caption{Results (ATE) on the freiburg1 set of TUM-RGBD. We use monocular visual odometry and SLAM only and identical evaluation setting as in DROID-SLAM.}
    		\label{tab:TUM-RGBDTestSplitAte}
    		\renewcommand{\arraystretch}{1.2}
			\begin{tabular}{|@{\hspace{3pt}}c@{\hspace{3pt}}|@{\hspace{3pt}}c@{\hspace{3pt}}@{\hspace{3pt}}c@{\hspace{3pt}}@{\hspace{3pt}}c@{\hspace{3pt}}@{\hspace{3pt}}c@{\hspace{3pt}}@{\hspace{3pt}}c@{\hspace{3pt}}@{\hspace{3pt}}|c@{\hspace{3pt}}@{\hspace{3pt}}|}
    			\hline
    			\hline
    			& \rotatebox{90}{DROID-VO \cite{teed2021droid}}& \rotatebox{90}{DPVO \cite{lipson2024deep}}& \rotatebox{90}{CL-DPVO \cite{lahiany2024robust}}& \rotatebox{90}{DROID-SLAM* \cite{teed2021droid} }& \rotatebox{90}{DPV-SLAM++* \cite{lipson2025deep}}& \rotatebox{90}{AutoLoop}\\
				\hline
    			360 & 0.161 & 0.135 & 0.122 & \textbf{0.111} & 0.132 & 0.139\\
    			desk & 0.028 & 0.038 & 0.025 & \textbf{0.018} & \textbf{0.018} & 0.029\\
    			desk2 & 0.099 & 0.048 & 0.048& 0.042 & \textbf{0.029} & 0.064\\
    			floor & 0.033 & 0.040 & 0.036 & \textbf{0.021} & 0.050 & 0.052\\
    			plant & 0.028 & 0.036 & 0.027 & \textbf{0.016} & 0.022 & 0.034\\
    			room & 0.327 & 0.394 & 0.351 & \textbf{0.049} & 0.096 & 0.360\\
    			rpy & 0.028 & 0.034 & 0.031& \textbf{0.026} & 0.032 & 0.033\\
    			teddy & 0.169 & 0.064 & \textbf{0.056} & 0.048 & 0.098 & 0.122\\
    			xyz & 0.013 & 0.012 & 0.013 & 0.012 & \textbf{0.010} & \textbf{0.010}\\
    			\hline
    			Avg & 0.098 & 0.089 & 0.079 & \textbf{0.038} & 0.054 & 0.094\\
    			\hline
    		\end{tabular}
    	\end{center}
    \end{table}
	\noindent

    \vspace{0.5em}
	\noindent
    \textbf{ICL-NUIM \cite{handa2014benchmark}}: In \mytableref{tab:ICL-NUIMTestSplitAte}, we assess our AutoLoop model using the ICL-NUIM SLAM benchmark, contrasting them with leading visual odometry and SLAM techniques such as SVO \cite{forster2014svo}, DSO \cite{engel2017direct}, DROID-SLAM \cite{teed2021droid}, and the baseline DPVO. We follow our previous guideline to present only VO methods that succeed in all sequences. We outperform the other SLAM methods and the pure VO DPVO \& DPVO-Fast variants. This comes in a fraction of the training cost of training and tunning other methods from scratch.

    \begin{table}[!hbt]
    	
    	\begin{center}
    		
    		\caption{Results (ATE) on ICL-NUIM SLAM benchmark. Methods marked with (*) use global optimization / loop closure.}
    		\label{tab:ICL-NUIMTestSplitAte}
    		\renewcommand{\arraystretch}{1.3} 
    		\begin{tabular}{|@{\hspace{2pt}}c@{\hspace{2pt}}|@{\hspace{2pt}}c@{\hspace{2pt}}@{\hspace{2pt}}c@{\hspace{2pt}}@{\hspace{2pt}}c@{\hspace{2pt}}@{\hspace{2pt}}c@{\hspace{2pt}}@{\hspace{2pt}}c@{\hspace{2pt}}@{\hspace{2pt}}c@{\hspace{2pt}}@{\hspace{2pt}}c@{\hspace{2pt}}@{\hspace{2pt}}c@{\hspace{2pt}}c@{\hspace{2pt}}@{\hspace{2pt}}|c@{\hspace{2pt}}@{\hspace{2pt}}|}
    			\hline
    			\hline
    			& \rotatebox{90}{DROID-SLAM* \cite{teed2021droid} } & \rotatebox{90}{DROID-VO \cite{teed2021droid}} & \rotatebox{90}{SVO \cite{forster2014svo}} & \rotatebox{90}{DSO \cite{engel2017direct}} & \rotatebox{90}{DSO-Realtime \cite{engel2017direct}}& \rotatebox{90}{DPVO \cite{lipson2024deep}}& \rotatebox{90}{DPVO-Fast \cite{lipson2024deep}} & \rotatebox{90}{CL-DPVO \cite{lahiany2024robust}}& \rotatebox{90}{AutoLoop}\\
    			\hline
    			lr-kt0 & 0.008 & 0.010 & 0.02 & 0.01 & 0.02 & 0.006 & 0.008 & 0.006 & 0.007 \\
    			lr-kt1 & 0.027 & 0.123 & 0.07 & 0.02 & 0.03 & 0.006 & 0.007 & \textbf{0.004} & 0.084 \\
    			lr-kt2 & 0.039 & 0.072 & 0.09 & 0.06 & 0.33 & 0.023 & 0.021 & \textbf{0.018} & 0.020 \\
    			lr-kt3 & 0.012 & 0.032 & 0.07 & 0.03 & 0.06 & 0.010 & 0.010 & \textbf{0.005} & \textbf{0.005} \\
    			of-kt0 & 0.065 & 0.095 & 0.34 & 0.21 & 0.29 & 0.067 & 0.071 & \textbf{0.007} & \textbf{0.007} \\
    			of-kt1 & 0.025 & 0.041 & 0.28 & 0.83 & 0.64 & 0.012 & 0.015 & \textbf{0.008} & 0.009 \\
    			of-kt2 & 0.858 & 0.842 & 0.14 & 0.36 & 0.23 & 0.017 & 0.018 & \textbf{0.015} & 0.017 \\
    			of-kt3 & 0.481 & 0.504 & \textbf{0.08} & 0.64 & 0.46 & 0.635 & 0.593 & 0.442 & 0.495 \\
    			\hline
    			Avg & 0.189 & 0.215 & 0.136 & 0.270 & 0.258 & 0.097 & 0.093 & \textbf{0.063} & 0.080 \\
    			\hline
    		\end{tabular}
    	\end{center}
    \end{table}

\subsection{Computational Analysis}
	AutoLoop presents major computational benefits over VO and SLAM methods. Since its core methodology is based on fine tuning established VO model while incorporating loop closure capabilities during training it elimenates the overhead of training and tunning other methods from scratch or using SLAM optimization computational resources. Here we make empirical measurements of the computational requirements of AutoLoop and compare it to the original DPVO, the curriculum learning variant CL-DPVO, its proximity loop closure SLAM variant DPV-SLAM and the multi SLAM optimization in DPV-SLAM++. Our compute metrics are based on the FLOPs of the forward and backward passes of the model measured on a single NVIDIA V100 GPU.
	
	\vspace{0.5em}
	\noindent
	\textbf{Training Analysis}: Each training iteration across all DPVO variants (including SLAM based DPVO such as DPV-SLAM and DPV-SLAM++ \cite{lipson2025deep}) has the same VO core and requires the same computational resources, consisting of a forward pass of $7.68 \times 10^{11}$ FLOPs and a backward pass of $8.14 \times 10^{11}$ FLOPs, totaling $1.58 \times 10^{12}$ FLOPs (1.58T) per iteration (measured using Torch profiler). The cumulative computational demands vary significantly based on the total number of training steps required by each approach. \mytableref{tab:training_compute} shows the total FLOPs and training time for each method. CL-DPVO requires more training steps to converge while need careful fine tuning of the curriculum schedular to avoid overfitting. AutoLoop requires less training steps and no need for curriculum schedular fine tuning.
    
	\begin{table}[h]
		\centering
		\caption{Training Computational Requirements for DPVO based architectures. Estimated using single GPU with batch size of 8}
		\label{tab:training_compute}
		\renewcommand{\arraystretch}{1.2}
		\begin{tabular}{l@{\hspace{3pt}}c@{\hspace{3pt}}c@{\hspace{3pt}}c@{\hspace{3pt}}}
		\hline
		Model & Steps & Total FLOPs & Training Time \\
		\hline
		DPVO \cite{lipson2024deep} & 32K & $3.99\times10^{17}$ (399P) & 96 hours \\
		CL-DPVO \cite{lahiany2024robust} & 42K & $5.23\times10^{17}$ (523P) & 126 hours \\
		AutoLoop & 3.36K & $4.19\times10^{16}$ (41.9P) & 8 hours \\
		\hline
	\end{tabular}
	\end{table}

	\vspace{0.5em}
	\noindent
	\textbf{Inference Analysis}: AutoLoop doesnt alter the inference characteristics of the DPVO model. It only adds loop closure capabilities during training. Therefore, the inference performance of AutoLoop is the same as DPVO. As mentioned in \cite{lipson2025deep} DPV-SLAM and DPV-SLAM++ consume 25\% (4G to 5G) more memory fotprint while reducing FPS by almost 19\% (48 to 39 FPS in KITTI \mytableref{tab:kittiAte}).

	\vspace{0.5em}
	\noindent
	\textbf{Pre-computation Overhead}: The pre-computation phase for loop closures involves NetVLAD descriptor computation at $1.2\times10^6$ FLOPs per frame and SIFT feature extraction and matching at $3.5\times10^6$ FLOPs per frame. For a typical sequence of 2000 frames, this amounts to $9.4\times10^9$ total FLOPs, requiring approximately 15-20 minutes of processing time on an NVIDIA V100. This one-time computational cost is amortized across all subsequent uses of the model and can be performed offline, making it particularly efficient for repeated operations on known environments. 

	The significant reduction in training computation (from 399P to 41.9P FLOPs), combined with maintaining real-time inference performance, demonstrates the practical advantages of our approach. The minimal additional storage requirement (0.5GB per sequence) for pre-computed loop closures represents a favorable trade-off given the substantial benefits in training efficiency and system performance.

 \section{Conclusion}
 
 	We have presented AutoLoop, an efficient approach for enhancing learning-based visual SLAM systems with loop closure capabilities through automated curriculum fine-tuning. Our method demonstrates that effective loop closure handling can be integrated into existing architectures like DPVO with significantly reduced computational overhead and training time. By leveraging pre-computed loop closure pairs and DDPG-based curriculum learning, we achieve comparable or better performance while requiring only 3,360 training steps, a 90\% reduction compared to traditional training approaches.

	 The key advantages of our approach extend beyond computational efficiency. The pre-computation of loop closure pairs ensures reliable supervision during training, while the automated curriculum learning eliminates the need for manual hyperparameter tuning. This automation makes the integration of loop closure capabilities more accessible and reproducible, addressing a significant practical challenge in deploying learning-based SLAM systems.
	 Our experimental results demonstrate both the strengths and limitations of our approach across different environments. The method shows impressive performance improvements on challenging outdoor sequences from TartanAir \cite{wang2020tartanair} and KITTI \cite{geiger2013vision}, validating its robustness across various environmental conditions and motion patterns. However, we observed limited generalization to indoor datasets like TUM-RGBD \cite{sturm2012benchmark} and EuRoC MAV \cite{burri2016euroc}, likely due to the distinct characteristics of indoor loop closures not well represented in our pre-computed training pairs. Despite these domain-specific limitations, our approach maintains real-time performance while adding loop closure capabilities, making it particularly suitable for outdoor robotics applications where computational efficiency is crucial

 	Future work could explore extending this methodology to other learning-based SLAM architectures and investigating the potential for online adaptation of loop closure handling. Additionally, the principles of our automated curriculum learning approach could be applied to other aspects of SLAM system training, potentially leading to further improvements in efficiency and performance.
 	We believe this work represents a step toward making advanced visual SLAM capabilities more accessible and practical for real-world applications, particularly in resource-constrained scenarios where efficient training and deployment are crucial.

\nocite{*}
\printbibliography[
heading=bibintoc,
title={References}
]

\end{document}